# Who's the Best Detective? LLMs vs. MLs in Detecting Incoherent Fourth Grade Math Answers

## Abstract

Written answers to open-ended questions can have a higher long-term effect on learning than multiple-choice questions. However, it is critical that teachers immediately review the answers, and ask to redo those that are incoherent. This can be a difficult task and can be time-consuming for teachers. A possible solution is to automate the detection of incoherent answers. One option is to automate the review with Large Language Models (LLM). In this paper, we analyze the responses of fourth graders in mathematics using three LLMs: GPT-3, BLOOM, and YOU. We used them with zero, one, two, three and four shots. We compared their performance with the results of various classifiers trained with Machine Learning (ML). We found that LLMs perform worse than MLs in detecting incoherent answers. The difficulty seems to reside in recursive questions that contain both questions and answers, and in responses from students with typical fourth-grader misspellings. Upon closer examination, we have found that the ChatGPT model faces the same challenges.

## Introduction

Written answers to open-ended questions could have a higher long-term effect on learning than multiple-choice questions. They help improve predictions of long-term learning as measured by end-of-year standardized tests (Urrutia et al, 2022). They challenge students to think critically, apply their knowledge, and explain their thinking. They are a type of desired difficulty. Initially, writing can complicate learning, but experimental studies have found that it improves retention and comprehension in the long term (Bicer et al, 2013). Some meta-analysis shows a significant positive effect size on student achievement of 0.42 SD (Bicer et al, 2018).

In addition, being able to not only solve math problems but also explain the solving process in writing is becoming an increasingly important skill in the mathematics curriculum. For example, the Common Core State Standards (CCSS) (National Governors, 2010) highlights the importance of writing and the need for coherent, well-constructed arguments. CCSS and other curriculum frameworks are highlighting the importance of combining language and mathematics to develop students' argumentative skills and ensure they are successful in their mathematics studies. The Task Force on Conceptualizing Elementary Mathematical Writing (Casa et al., 2016) emphasizes the importance of language in mathematics lessons. In Germany, language competencies in mathematics lessons are gaining more attention (Ruwisch et al, 2014).

However, the complex thinking required when explaining the reasoning behind a mathematical problem is a big challenge for fourth graders. In order to provide an effective written argument, students have to be able to explain their understanding of the problem, their strategy for solving the problem, and their results. Fourth-graders are still developing their math skills and may not have the language skills needed to explain their thoughts and reasoning. They struggle to identify and explain the evidence and reasoning behind their problem-solving strategies.

Writing is a complex process. Therefore, it is very helpful that the teacher is aware in real time of what their students are writing. In this way the teacher can give them timely and real-time feedback. It is also very important to detect students with misbehaviors and who may be writing answers that are incoherent. This is different from incorrect answers. Incoherent answers denote a negative attitude. Most probably, the student does not want to answer. The teacher has to detect these negative attitudes early. Thus, teachers could immediately review the answers, and ask to redo those that are incoherent. Therefore, it is important that the teacher detect this situation and try to control it. However, with 20 or more fourth-graders in the classroom, it is very demanding that the teacher review all answers and send feedback in real-time to each student. This is a difficult task and can be time-consuming. A solution is to automate the detection of incoherent answers. This, in principle, is possible if students are writing on an online platform. Moreover, if students are writing comments to peers in a peer-review activity, it is also helpful that the teacher review the comments. It is crucial to prevent negative attitudes from spreading. They can spread the moods to other students and hinder the climate and management of the class. This review also takes time and is very demanding for the teacher.

The detection of incoherent responses is also a very prominent factor for the development of chats that are intelligent interactive tutors. If the chat cannot detect incoherent responses, then its performance as a tutor will be severely limited. In our experience, with typical responses from fourth-graders from hundreds of vulnerable schools, about 20% of the responses are incoherent. They reveal a possible negative attitude of the student. This is different from incorrect answers. An intelligent tutor's strategy for incoherent answers should be different from that for incorrect answers.

In order to automate the detection of incoherent answers, there are two possibilities. One possibility is to automate the review with LLM. Another possibility is to automate with Machine Learning (ML) (Urrutia et al, 2023). Since ML requires extensive training, it is interesting to explore the use of LLM.

LLMs do not require training and have demonstrated several characteristics of sophisticated cognitive processes. For example, they have incredible writing and summarizing skills. They also have social reasoning skills. In this regard, they are equivalent to 9-years old. ChatGPT passes the Theory of Mind (ToM) tests for 9-year-old students (Kosinski, 2023). This means that ChatGPT can "guess" what is going on in another person's mind based on available text information. Additionally, (Binz et al, 2022) tested GPT-3 and found it performs reasonably well on several tests on decision-making, information search, deliberation, and causal reasoning tasks. Furthermore, (Schwitzgebel, 2023) found that when faced with complex philosophical questions, it can be very difficult to distinguish answers written by a fine-tuned GPT-3 model that emulates an author from those answers written by the emulated philosopher. This recognition task is difficult even for professional philosophers who are specialists in the emulated author.

However, LLMs reasoning abilities are still limited. For example, when asked to consider the request: "If my hand is closed then I have a coin. What can be concluded if I do not actually have a coin?" OpenAI playground incorrectly responds: "If you really don't have a coin, then you can conclude that the original statement is not correct." LLMs, such as OpenAI playground, also do not pass the Wason Selection task. This is a classic abstract reasoning test. It is a difficult test, that less than 25% of human adults do it correctly. However, there are non-abstract versions. One interesting version is a social contract formulation (Cosmides et al., 1992). This version is much easier for humans than the abstract one and it is also easier than other non-abstract versions.

Imagine that you have to enforce the following law: "If a person is drinking beer, then he must be over 20 years old". In addition, imagine that there are four people sitting at a table in a bar. Person 1 is drinking beer. Person 2 is drinking coke. Person 3 is 25 years old. Person 4 is 16 years old. Indicate only those persons you need to check information to see if any of these people is breaking the law. The answer of OpenAI playground is wrong: Person 1 and Person 3.

In contrast to the abstract version, 75% of human subjects answer correctly the social contract formulation (Cosmides et al, 1992). Even though the logic is the same in both formulations, people unconsciously and effortlessly use a different reasoning mechanism to solve it. It is a mechanism of completely different nature: a cheater detection one (Cosmides et al, 1992).

According to (Huang et al, 2022) it is still unclear to what extent LLMs can do true reasoning. (Bang et al, 2023) found that ChatGPT is 64.33% accurate on average in 10 different reasoning tasks. They conclude that it is an unreliable reasoner. Similarly, (Mialon et al, 2023) concludes that meaningful augmentations in fully self-supervised language models (LMs) is still an open research question. (Frieder et al, 2023) tested the ChatGPT's mathematical abilities. They found that the capabilities are significantly below those of an average mathematics graduate student. Furthermore, specialized LLMs trained on math papers, like Minerva, make mistakes with big numbers (Castelvecchi, 2023).

In this paper, our research questions are:

- How do LLMs fare in detecting incoherent fourth-graders responses to typical math word problems?
- How does the incoherence detection performance of LLMs compare to that of ML classifiers?

To address these questions it is important to analyze how LLMs do it by reviewing answers to typical questions posed by teachers. These should not be invented questions for a research study. That is, we need to use a representative sample of word problems that teachers have actually already placed in their fourth grade classes. In addition, to ensure representativeness, we need, on the one hand, to have a variety of teachers. Therefore, we have to make sure that we have questions posted by several different teachers. On the other hand, we have to measure the performance of LLMs in reviewing the actual answers of all the fourth graders of those teachers. Given the excellent discursive capacity of LLMs, it is very interesting to investigate how well they detect incoherent responses and the explanations they provide.

## Related work

First, we consider the use of Machine Learning (ML) to detect incoherent written math answers to open-ended questions. There is an extensive literature in automated short answer grading (ASAG). (Wang et al., 2021) found that automated scoring systems with simple hand-feature extraction can accurately assess the coherence of written responses to open-ended questions. The study revealed that a training sample of 800 or more human-scored student responses per question was necessary to construct accurate scoring models. However, this requirement of hundreds of responses labeled per question makes it not practical for most classrooms.

In contrast, in (Urrutia et al., 2023) we analyzed fourth graders' written answers to open-ended questions by fourth-graders. In this case, there were between 20 to 60 answers per question. We tested the classifier in a completely different set of questions and students. Using Machine Learning, we trained question and answer classifiers. We found that the classifier achieves 79.15% F1-score for incoherent detection. Moreover, in (Urrutia et al., 2022) we found that the classifiers help improve predictions of end-of-year national standardized tests results.

Second, let's consider the use of LLM. There is an increasing number of studies on the impact and opportunities of using LLMs in education (Zhai, 2022; Kasneci et al, 2023). For example, (Jalil et al., 2023) analyzes the performance of ChatGPT when solving common questions in a dataset that contains questions from a well-known software-testing book. They found it can provide correct answers in 44% of the cases, and partially correct in 57% of the cases.

(Kasneci et al., 2023) examined opportunities for elementary school students, middle and high school students, university students, group and remote learning, learners with disabilities and professional training. They conclude that despite certain difficulties and challenges these models offer many opportunities to enhance students´ learning experience and support the work of teachers.

A common type of word problems in elementary school mathematics are those that require several steps of reasoning. It is therefore important to look at those capabilities of the LLMs.

According to (Kojima et al, 2023) the performance of LLMs in intuitive and single-step reasoning tasks is excellent using task-specific few-shot or zero-shot prompting (Liu et al., 2021). These are tasks that cognitive psychologists identify as system-1 mental procedures (Stanovich and West, 2000; Stanovich et al, 2016), However, in cognitively more sophisticated tasks the performance is much lower. These tasks require system-2 capabilities. They may require the inhibition of system-1 mental procedures, and several other processes. For example, seeking various points of view before concluding, and weighing the advantages and disadvantages before deciding.

(Rae et al., 2022) studies the performance of a Gopher, a LLM with up to 280 billion parameters. They conclude that the performance is weak on tasks requiring multi-step reasoning like in math and logical reasoning. Moreover, they conclude that in certain math and logical reasoning tasks, it is unlikely that more parameters will improve performance.

To overcome this difficulty, there are recent attempts to improve reasoning performance of LLMs using prompt engineering. Prompt-based learning modifies the original input using a template into a text string message that has some empty spaces, and then the LM fills in the empty information to get a final string (Liu et al, 2017). With it, the user obtains the result. Selecting the appropriate indications is very important. The idea is to manipulate the behavior of the model so that the pre-trained LLM itself predicts the desired outcome. Sometimes, there is no need for additional task-specific training.

Another option is the use of Chains of Thought (CoT). (Kojima et al., 2023) use it to solve elementary math word problems. This is the case where the user presents a word problem but it can easily confuse ChatGPT. The CoT strategy is to tell ChatGPT to do it step-by-step, and then use the response to ask to extract the numerical result. (Kojima et al., 2023) found that CoT strategy generates a significant performance improvement.

In the word problems that teachers have been giving fourth graders over the years, a large proportion are questions that contain within themselves a question. In addition, the question also contains a character that gives an answer. Therefore, the student's problem is to comment on the character's answer. The student has to decide if it is correct or incorrect, and give a written explanation. Thus, there is not only a multi-step reasoning problem but also one of recursion. This means, a constituent that contains a constituent of the same kind (Pinker et al, 2005). The student must comment on a character's answer to a mathematical question that already requires multi-step reasoning. Additionally, in the peer review stage, each student must also comment on the comments that a classmate gave to one of those problems. That is, there is a recursion of even higher order.

Therefore, we believe that our paper is the first research on the use of LLMs to analyze this type of students' answers. Since these are very common types of word problems, we need to support the teacher by automating the online review of incoherence in students´ written answers to this type of open questions. This is particularly important for elementary school students who are just beginning to argue in written form. In the classroom environment, it is necessary to determine immediately the coherence of the answers. This allows the teacher to control negative attitudes and request students to redo incoherent answers. This is important for the classroom climate.

In this paper, we compare the performance of LLMs with ML classifiers in a reasonably big number of real world fourth graders' written responses to open-ended math questions.

# Materials and methods

The materials employed in this study consist of questions and answers generated through the ConectaIdeas online platform (Araya et al., 2015; Araya et al., 2019; Araya Diaz, 2020). The questions are spontaneously created by teachers and are subsequently answered by fourth-grade students during weekly math exercise sessions, which typically last 90 minutes

(Urrutia et al., 2023). The platform asks open questions and closed questions, such as essay questions and multiple-choice questions, respectively. We only use open-ended questions, which provide unstructured written text responses.

In contrast to other Educational Data Mining (EDM) tasks, such as Automatic Short Answer Grading (ASAG), this study focuses on identifying incoherence in responses to open-ended questions (Urrutia et al., 2023). The detection of incoherence in the open-ended responses is achieved through the application of Generative Pre-trained Transformers (GPT), which is a novel family of Natural Language Processing (NLP) techniques.

Our methodology involves collecting question-answer pairs, using various GPT models and prompting strategies to develop classifiers, and subsequently assessing their performance on a test set in comparison with ML models. Figure 1 illustrates this process.

*Figure 1. The four LLMs used in this study. Three of them with prompting of zero, one, two or three shots. Below are the two databases with questions and answers. Those of 2019 will be used to train the ML models. The 2017 one is used for testing.*

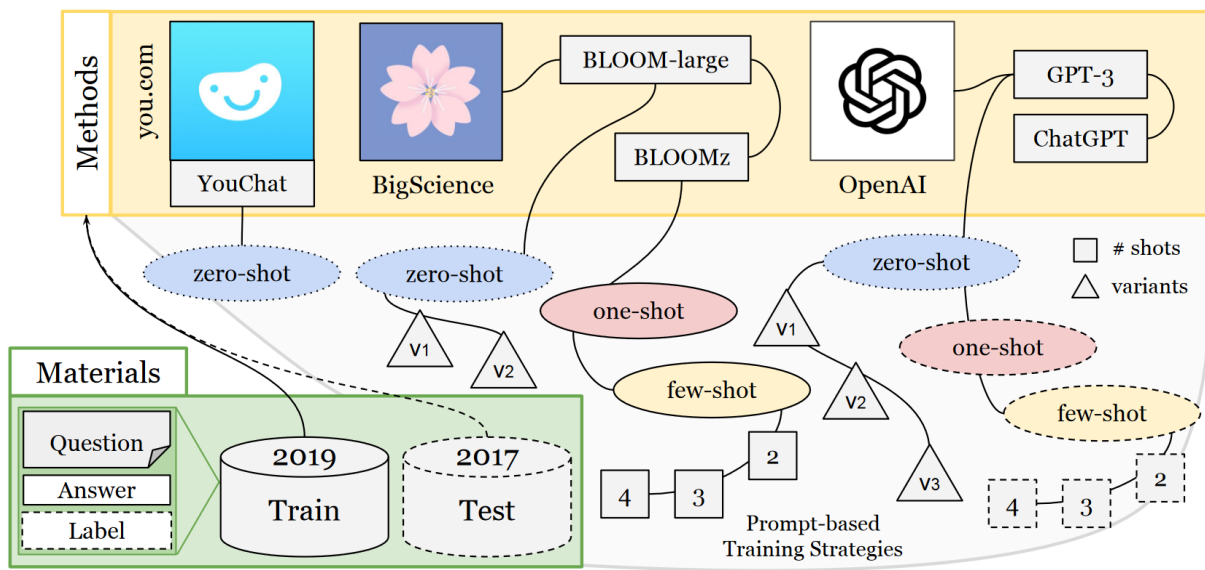

## Materials

The dataset comprises responses to open-ended questions during 90 minutes fourth grade sessions in 2017 and 2019. All the question-answer pairs were posed by 12 teachers, answered by 974 fourth graders, and then labeled by 10 teachers. The questions varied in nature, with some being conceptual, while others revolved around specific situations and characters. Moreover, certain questions were explicitly designed to elicit explanations.

On the ConectaIdeas platform, teachers posed six types of open-ended questions (Urrutia et al, 2023). Q1-questions involve calculating a quantity without providing an explanation or justification for the answer. Q2-questions, require both calculation of a quantity and a written explanation or justification. Q3-questions introduce characters and statements, and may involve deciding who is right or evaluating the correctness of a character's statement. In both cases, the question demands a justification. Q4-questions ask for a comparison of two quantities, with the task of identifying which quantity is greater or explaining why quantities are equal or different. Q5-questions require to write a problem asking content-related questions. Finally, Q0-questions are those that do not fit into any of the previous categories.

The linguistic characteristics of students' responses are influenced by the type of question. Depending on the question, students may need to provide integers, decimals, fractions, character names, explanations, or a simple "yes" or "no" response. The study identifies two types of incoherent answers (Urrutia et al, 2023). The first type of incoherence is question-independent incoherent answers. This includes responses that contain emoticons, laughter, curse words, phonemic errors, missing letters, consonant substitutions, and repeated words. The second type of incoherence are question-dependent incoherent answers. This occurs when the question is crucial in determining whether the answer is coherent or not. For instance, answering "no" to a type Q1 question may be considered incoherent.

*Table 1. Distribution of questions and answers of the sample of the 2017 data (test set) by question type.*

|  | Type of question | | | | | |
|---|---|---|---|---|---|---|
|  | **Q0** | **Q1** | **Q2** | **Q3** | **Q4** | **Q5** |
| **Coherent** | 107 | 8 | 39 | 262 | 11 | 114 |
| **Incoherent** | 42 | 1 | 18 | 41 | 5 | 29 |
| **Questions** | 68 | 5 | 35 | 161 | 9 | 82 |

The test data set are all questions posed by teachers in 2017. They were not used to train the ML classifiers. Ten teachers labeled the answers. We selected only answers with agreement between the two teachers that labeled them. The dataset contains 360 questions and 677 answers. Table 1 describes the number of questions types, and the number of answers labeled coherent and incoherent for each type of question. The most frequent type of question is type Q3. We call recursive this type of questions, since the question contains a question and an answer. The student has to comment about the accuracy of the answer.

In the present work we added the label of presence of spelling errors in all the answers of the year 2017. In addition, we manually classified all questions according to the six question types. This addition of new labels will later allow us to perform a detailed study of the predictive performance of the GPT model by question type and whether the answer has child misspelling or not.

# Methodology

## Objective and Experiments

The main goal of this study is to evaluate the effectiveness of GPT models in detecting incoherences in open-ended responses. Given the popularity of GPT models in NLP and text generation, it is important to determine their suitability in identifying incoherent answers from fourth-grade students. To accomplish this, it is essential to select appropriate GPT models that support the language of the dataset. Therefore, we will assess a few GPT models that meet our selection criteria and evaluate them using various prompting strategies, including zero-, one-, and few-shot learning.

Additionally, we will employ ML models to gain insights into the challenges faced by GPT models. To comprehensively evaluate the GPT models, we will leverage ChatGPT to examine their performance intricacies and assess their ability to explain their predictions. Our study will focus on identifying the key obstacles associated with GPT models' ability to identify incoherences in fourth-grade students' open-ended math responses.

## Task Description

To automatically detect incoherence in answers to open-ended questions, we will use both the question and the answer. By "incoherence", we mean answers that contain irrelevant or unrelated information, lack coherence, or fail to address the question being asked. It's important to note that not all incorrect answers are incoherent, and incoherence can take various forms.

Incoherence can manifest itself in several ways, such as the presence of illegible text, emoticons, or laughter, but it's not limited to these examples. For instance, answers with random letters or the presence of cursed-words are also indicators of incoherence.

To identify incoherent answers, we will use a binary task with labels "Coherent" and "Incoherent". Our first work Urrutia et al, (2023) suggests that incoherence is highly dependent on the type of question being asked. For example, answers to question type Q1 are not coherent if they do not have appropriate numerical representations. While answers to question type Q2 will be coherent if they have numbers and an explanation of the mathematical expression.

## Step-By-Step Route

We intend to assess the predictive performance of various GPT models by subjecting them to different experiments, using various strategies. However, limitations such as budget constraints, availability, and privacy, among other factors, make it challenging to explore the full potential of GPT models comprehensively (Bender, E. M. et al, 2021). Therefore, we opted to explore selected GPT models in-depth, within their limitations, using various routes. Our optimal route is depicted in Table 2, which we developed to compare the individual characteristics of each model and determine the best way to exploit their architectures and strategies via extensive experiments.

*Table 2. Comparative table of characteristics. Note: Yes (Y) and No (N).*

|  | Open-source | Interface | API | Privacy | Budget constraints | Usage limits |
|---|---|---|---|---|---|---|
| **YouChat** You.com | N | Y | N | Y | N | Y |
| **ChatGPT** OpenAI | N | Y | N | N | N | N |
| **GPT-3** (text-davinci-003) | N | Y | Y | Y ? | Y (*first usage free) | N |

|  | Open-source | Interface | API | Privacy | Budget constraints | Usage limits |
|---|---|---|---|---|---|---|
| **BLOOM-large** BigScience | Y | N | Y | Y | Y (*increase ratio) | Y |
| **BLOOMz** bloomz-560m | Y | N | Y | Y | N | N |

Our methodology involves using the YouChat model as the baseline model for the GPT models, with the zero-shot strategy. YouChat is a GPT model that is adjusted to be conversational and is free. We found that it performs well in classifying responses according to coherence with only one task description, and it is based on GPT-3.5. Additionally, the YouChat model's features enable us to study the best way to make prompts. Although the YouChat model has no API to perform a simple task and its inference time on the entire test set is considerably time-consuming compared to other GPT models, we decided to study it using only the zero-shot strategy due to its unique features.

We only used the ChatGPT model on a small subset of the test data due to the lack of API support at the time of writing. Despite the high demand for the model, it is available free of charge, and OpenAI has provided documentation and tutorials for its use. However, the pre-trained weights and model-specific source code are not currently available in a public repository on GitHub. Hence, the only way to use the model is through the interface.

The BLOOMz model is a small version of the BLOOM model that can be downloaded and performed locally. This model is trained to be used with prompting, unlike the original BLOOM model, and supports Spanish language. Moreover, it is open-source, making it an ideal model to use with various strategies. However, we decided not to use Petals, a version of BLOOM that allows simple inference, due to computational and memory resource limitations. In addition, it is necessary to pay to increase the inference rate, making it not feasible for our study. On the other hand, the GPT-3 model is paid but offers a free and reasonable first period for testing two experiments. As a result, it is the basis for several other GPT models such as ChatGPT.

We will follow the following steps in our study: First, we will use YouChat to study the incoherence detection task's prompting. Second, we will use it with the zero-shot strategy as the baseline model of the GPT models. Third, we will use BLOOM-large with only the zero-shot strategy. Fourth, we will study the GPT-3 and BLOOMz models with the one- and few-shot strategies. Fifth, we will test the GPT-3 model with the zero-shot strategy. Finally, we will use ChatGPT to study the shortcomings and challenges of the GPT models.

## Evaluation

We adopted a specific data split strategy for the training and testing of our ML classifiers. Specifically, we utilized the 2019 dataset for training purposes and the 2017 dataset for testing. This approach was motivated by the desire to assess the model's performance in a real-world scenario, where it is exposed to a different set of students and questions. Additionally, by using a dataset from a different year, we could evaluate the model's generalizability and capacity to perform well on novel data.

The performance of the ML classifiers was evaluated using a data splitting strategy, where the dataset was divided into training and test sets. The model's performance was compared against two baseline models, a rule-based unsupervised model and XGBoost+Mix, and two ML models previously trained by Urrutia et al. (2023), a general XGBoost model, and BETO-mt, a deep learning model based on the Spanish version of BERT (Devlin, J. et al., 2018), called BETO (Cañete, J. et al. 2020). The evaluation approach utilized three metrics beyond traditional accuracy measurement: Precision, Recall, and F1-score. The test set consisted of 541 coherent pairs and 136 incoherent pairs, and the size of the test set was determined based on the available data.

The performance metrics chosen were based on the significant imbalance in the incoherent responses, which represent 13% of the total answers. The Precision metric was calculated as the proportion of the answers correctly predicted as incoherent over the total of answers predicted as incoherent, while Recall was defined as the proportion of the incoherent answers correctly predicted as such over the total of incoherent answers. The F1-score, which is the harmonic mean between Precision and Recall, was computed for both the incoherent and coherent classes. Additionally, we also track the Support metric, representing the number of occurrences of each particular class in the ground truth.

# Results

## GPT models' performance

In this section, we report on the performance of several GPT models, specifically YouChat, GPT-3, and BLOOM, utilizing zero-, one-, and few-shot strategies. We establish YouChat as the baseline model for GPTs using the zero-shot strategy, while GPT-3 and BLOOM are proposed models. We assess GPT-3's performance using three strategies: zero-shot, one-shot, and few-shot. Additionally, we evaluate the performance of BLOOM models in both its large (BLOOM-large) and smaller (BLOOMz) versions using zero-shot strategy for BLOOM-large and one- and few-shot strategies for BLOOMz. F1-score metrics were used to present the results of all experiments, as illustrated in Figure 3.

*Figure 3. Comparison of performance between GPT and ML models. (**Left**) F1-score of the Coherent class. (**Right**) F1-score of the Incoherent class.*

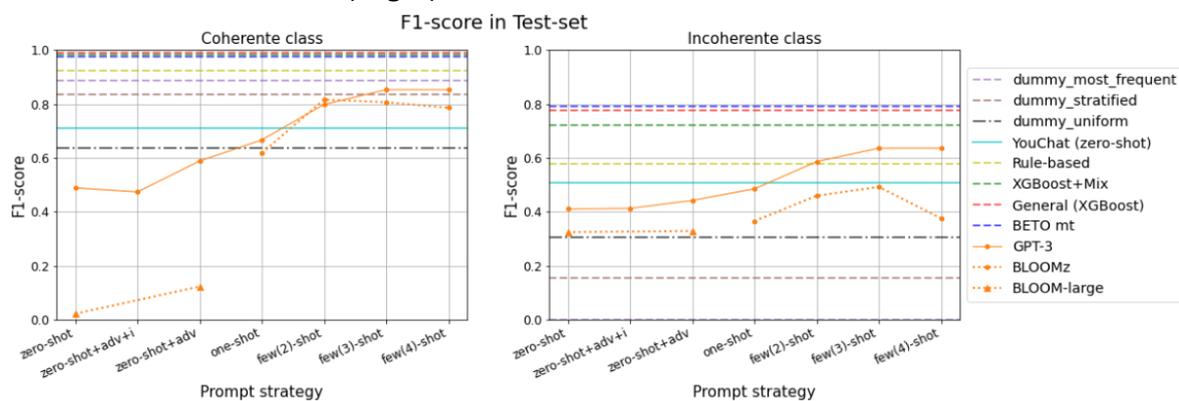

## YouChat

For our baseline, we evaluate the performance of four models for incoherence prediction: one strong model named YouChat, and three dummy models, including the Most frequent, Uniform, and Stratified classifiers (Urrutia et al., 2023). The dummy models do not consider the question-answer pairs and predict the class randomly. As reported in (Urrutia et al., 2023), the Most frequent classifier predicts only the majority class, resulting in low performance. The Stratified classifier predicts fewer answers as incoherent, resulting in lower precision than recall. The Uniform model achieves a better F1-score of 30% on the test set, and is used as a baseline for evaluating the performance of GPT models.

Figure 3 shows that the YouChat model with the zero-shot strategy outperforms the best dummy model, achieving a 50% F1-score in the incoherent class and a 70% F1-score in the coherent class. However, the YouChat model fails to outperform any of the ML models, which are trained with thousands of examples to accurately detect incoherence (Urrutia et al., 2023). The YouChat model is only introduced with a brief task description, and it is not trained to detect other features that may mislead the incoherence detection. This preliminary result suggests the need for further study on how to contextualize the models with key examples on coherence and incoherence in answers to open-ended questions.

## BLOOM

We evaluate the performance of the BLOOM model, consisting of the BLOOM-large and BLOOMz versions, using various prompting strategies. For the BLOOM-large model, we consider two zero-shot variants, the first including only a task description and prompt, and the second containing two additional key instructions related to correctness and recursion. We evaluate the BLOOMz model with prompting strategies that include task description, instructions, and examples from the training set, with one-shot and few-shot strategies (two to four examples) tested.

Figure 3 shows that the BLOOM-large model with the zero-shot variants performed poorly, achieving similar or worse performance than the random uniform model for both coherence and incoherence prediction. However, incorporating the two key instructions improved performance slightly. The BLOOMz model with the few-shot strategy, using three shots, achieved the best performance, with an F1-score of 49.19%, slightly lower than the 50% F1-score obtained with YouChat. The performance was lower than that of all ML models, suggesting that incorporating examples in the prompt helps to some extent, but the choice of appropriate examples and prompts may also be crucial. Further research is needed to explore the use of in-sample shots (i.e. examples in the test set) and additional information for incoherence detection.

## GPT-3

We analyze the performance of the GPT-3 model, using three training strategies: zero-, one-, and few-shot. The one-shot and few-shot strategies are examined with two to four shots, using in-sample shots, which are examples that are in the same set where we will test the model. In contrast, we study the zero-shot strategy with three variants, including the same variant as those used with the YouChat and BLOOM models. The second zero-shot variant involves a task description with two key instructions (correctness and recursion) and extra information from the incoherence detection task, and the third variant is the same as the second but without the extra information.

The GPT-3 model with the three variants of the zero-shot strategy performs considerably better than the BLOOM model, but not as well as the YouChat model. Incorporating more details in the task description improves the model's performance, but these additional details should be moderated. However, when we use the one-shot and few-shot strategies with the GPT-3 model, we obtain more encouraging results. Specifically, when we use three shots, the model outperforms the unsupervised rule-based ML model, achieving a 63.59% F1-score for incoherence prediction and 85.27% for coherence prediction. Although the GPT-3 model outperforms the YouChat and BLOOM models, it cannot outperform other ML models. Increasing the number of shots considerably improves the performance of the model, but the improvement is marginal when going from three to four shots.

The GPT-3 model with the few-shots strategy outperforms the YouChat and BLOOM models, in all their variants, but this result may be due to the fact that the shots used belong to the test set. The comparison of the three GPT models reveals that using generative models with zero-, one-, and few-shot strategies is not enough to compete against supervised ML models trained with thousands of examples. However, since prompts are a key part of GPTs model training, and the study only used a few prompts to evaluate the performance of these models, this conclusion may be equivocal. Future research will investigate what is happening with these models that are uncompetitive to refine the prompting and select more appropriate examples.

## Challenges encountered

We evaluated the performance of the GPT-3 model with a few-shot strategy in predicting incoherence in answers to open-ended questions. Despite achieving an F1-score of 63.59%, the GPT-3 model failed to outperform the best ML model reported in Urrutia et al. (2023). Specifically, Urrutia et al. (2023) found that the best ML model was the Spanish version of BERT called BETO-mt, which achieved higher precision in classifying coherent and incoherent answers. To determine why the GPT-3 model underperformed, we conducted further experiments and identified two challenges: Recursivity and Kids' Misspelling.

Table 2. Predictive performance of GPT-3 model and BETO-mt model. Note: Precision (P), Recall (R), F1-score (F); Support: 541 (Coherent) and 136 (Incoherent).

|          | Coherent |       |       | Incoherent |       |       |
|----------|----------|-------|-------|------------|-------|-------|
|          | P        | R     | F     | P          | R     | F     |
| **GPT-3**    | 97.16    | 75.97 | 85.27 | 48.82      | 91.18 | 63.59 |
| **BETO-mt**  | 95.47    | 93.53 | 94.49 | 76.19      | 82.35 | 79.15 |

Table 2 indicates that the GPT-3 model had a high recall of 91.18%, but low precision of 48.82%, which was significantly lower than the precision of the BETO-mt model (76.19%). These findings suggest that the GPT-3 model is more likely to classify coherent answers as incoherent, resulting in a large number of false positives. While the BETO-mt model had slightly lower recall, it achieved higher precision, suggesting it is less likely to make errors when classifying coherent answers as incoherent. Our study highlights the challenges associated with using generative language models like GPT-3 for predicting incoherence in open-ended answers and the potential benefits of using task-specific models like BETO-mt.

## Recursivity

We examined the challenges encountered in type Q3 questions, also known as choice and/or affirmation questions, which require the identification of correct statements made by characters within a given context. Specifically, we explored the impact of recursivity in open-ended questions, where a question and its answer are nested within the main open-ended question.

*Figure 4. Differences between Precision for the Incoherent class. (**Diagonal**) Difference of Precision between GPT-3 and BETO-mt models for each question type. (**Under diagonal**) Differences of Precision between questions for the GPT-3 model. (**Over diagonal**) Differences of Precision between questions for the GPT-3 model and BETO-mt model.*

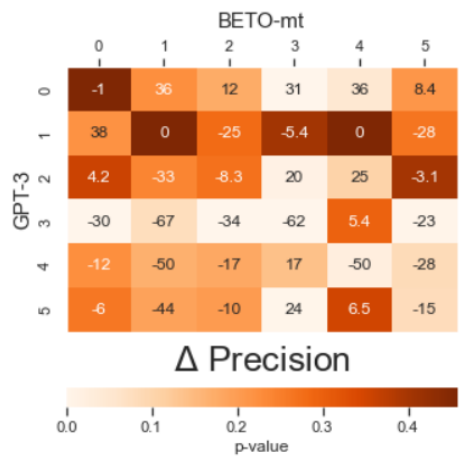

Recursivity poses a significant challenge for the GPT-3 model, as demonstrated by our precision comparison with the BETO-mt model in Figure 4. While the BETO-mt model achieved a precision of 94.6% for recursive questions, the GPT-3 model struggled significantly with an accuracy 62% lower than that of the BETO-mt model for Type Q3 questions. Moreover, the GPT-3 model performed worse on Q3-type questions than on other types of questions, except for Q2 and Q0 types, while the BETO-mt model demonstrated consistent precision across question types.

Overall, the lower performance of the GPT-3 model in type Q3 questions can be attributed primarily to the presence of recursivity in the open-ended questions. However, it is important to note that while the BETO-mt model exhibited higher precision, it was not perfect, and errors were observed even in type Q3 questions.

We encountered a challenge with the GPT model's ability to detect incoherence in open-ended questions of type Q3. To address this, we utilized the ChatGPT model, which is a similar architecture to GPT-3 but with fewer parameters and adjusted for natural language conversations with humans. Specifically, we focused on studying the behavior of ChatGPT on question-answer pairs with Recursivity, for which the GPT-3 model failed. Among the numerous examples we examined with the ChatGPT model, we present two relevant examples that highlight the difficulty of Recursivity.

In Example 3, a farmer needs to distribute 35 corn in seven trucks and prefers to ask a worker what amount to put in each truck. The worker suggests four corn per truck, but the correct answer is five corn per truck. The student correctly identifies the worker's answer as incorrect and justifies their answer by explaining that five corn in seven trucks equals 35 corn. However, when asked if the student's answer is coherent with the question, the ChatGPT model incorrectly identifies the answer as incoherent and provides a meaningless explanation. This may be due to the model confusing the student's answer with the worker's answer or the open-ended question with the question within it.

In Example 4, two friends buy different amounts of elastic, and a friend helps them calculate the total amount purchased. The friend suggests they bought 3 meters of elastic, but the correct answer is 3.32 centimeters. The student identifies the friend's answer as incorrect and explains that the correct amount is 3.32 centimeters. However, the ChatGPT model again identifies the answer as incoherent and provides an irrelevant explanation that fails to determine if the answer is coherent or not. The model may be confusing the coherence of the student's answer with the coherence of the friend's answer.

Our study highlights that Recursivity, the challenge of distinguishing between the open-ended question and the question within it, causes the ChatGPT model to fail to predict the coherence of the student's answer. We suggest studying this limitation in future applications of NLP.

Upon discovering the ambiguity produced by the original question, we recognized the necessity for a clear and unambiguous design of the prompt to prevent potential errors in the GPT-3 model's responses. To address this, we conducted an additional experiment utilizing a zero-shot prompt specifically tailored to Q3 type questions. This prompt design includes separate sections for the question and answer, with the answer labeled as "Felipe's answer" to differentiate it from the open-ended question's response. To minimize confusion, we asked the model three yes/no queries: (1) to evaluate the accuracy of the answer to the open-ended question, (2) to evaluate the accuracy of Felipe's answer (i.e., the student's response), and (3) to assess the coherence of Felipe's answer with the open-ended question.

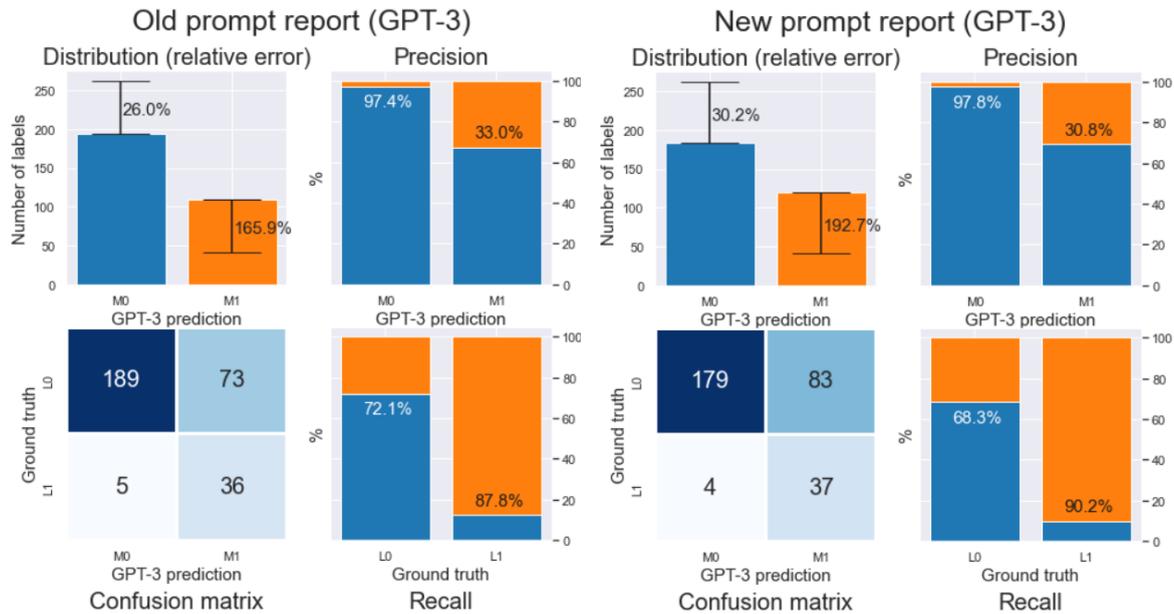

*Figure 5. Comparison of results of the GPT-3 model in question type Q3. (**Left**) GPT-3 with 4 shots. (**Right**) GPT-3 with new three-question zero-shot prompt.*

Despite this improved prompt design, as shown in Figure 5, the GPT-3 model's performance did not improve and, in fact, worsened. This suggests that the problem with Recursivity may stem from an internal feature rather than solely from prompt design. Thus, addressing Recursivity may require a more fundamental rethinking of the GPT-3 model. Furthermore, additional challenges could arise when analyzing recursive structures in other contexts, such as peer reviews where the model has to interpret comments from classmates on the student's response. It is also possible that the phrasing of the recursion statement may require revision to improve the performance of the GPT-3 model. Such efforts could have the complementary goal of enhancing the performance of the best ML models.

## Kids' misspelling

Fourth-grade students frequently make spelling errors, such as omitting punctuations, when answering open-ended questions in writing. However, these spelling errors differ from those of adults, as children tend to glue words together, break words into smaller units, or write words based on how they sound. In addition, students may use non-standard notations to write mathematical expressions. These types of childish misspellings occur frequently in responses to all open-ended questions. Responses with numerous spelling errors that differ significantly from adult misspelling can complicate analysis. Nevertheless, teachers do not consider responses with at least one spelling error as incoherent since such misspellings are common in children's writing. Consequently, responses with spelling errors are not necessarily incoherent or incorrect.

*Figure 6. Differences between Precision for the Incoherent class. (**Diagonal**) Difference of Precision between GPT-3 with and with-out misspellings for each question type. (**Under diagonal**) Differences of Precision between questions for the GPT-3 model in answers with misspellings. (**Over diagonal**) Differences of Precision between questions for the GPT-3 model in answers with-out misspellings.*

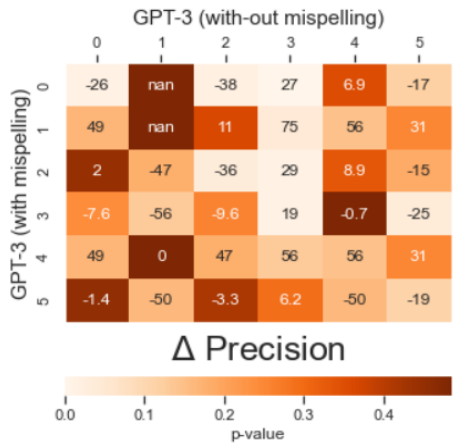

In Figure 6, we present a comparison of the precision obtained with the GPT-3 model when evaluated on responses with spelling errors and those without. We estimated the precision for each type of question to gain a better understanding of the effect of child spelling on each question type. Our results show that the GPT-3 model performs significantly better on responses without child spelling errors, except for questions type Q1 and Q4, which have few responses, where no statistically significant difference is observed. Conversely, for question type Q3, the effect of child spelling on precision is the opposite. However, we conclude that the problem in question type Q3 is not solely due to spelling errors but is related to recursion. Our findings demonstrate that child misspelling has a significant negative effect on the precision of the GPT-3 model for incoherence prediction.

Example 1 involves an open-ended knowledge question about tens, in which the student answers correctly but with a spelling error where two words are pasted together. Despite this error, teachers consider the answer coherent and correct. However, the ChatGPT model predicts the response as incoherent and nonsensical due to the spelling error. Additionally, the model provides an explanation that the pasted word is insufficient to answer the question, despite the information provided being the key to answering it.

Our results show that both models struggle with child spelling errors and that there is a limitation in their training domain. These findings suggest the need for further research to improve the models' ability to handle child spelling errors in open-ended questions.

## Discussion

The use of supervised learning has been the prevailing method in artificial intelligence (AI), particularly in ML and NLP (Liu, P. et al, 2023). However, recent developments have led to a shift away from hand-crafted features in favor of deep learning models that rely on learned and sometimes unsupervised representations (Haller, S. et al. 2022). This new approach involves designing a model with a fixed architecture that is pre-trained on large amounts of unstructured textual data using non-supervised tasks (Devlin, J. et al, 2018). The pre-trained LM is then fine-tuned for a specific task by updating its parameters based on the objective function. Urrutia et al. (2023) have explored a range of models, from ML models with hand-crafted features to fine-tuned LMs, for detecting incoherent answers to open-ended questions. The study found that models fine-tuned with BERT using a multi-task strategy outperformed traditional NLP models. However, a new paradigm has emerged in recent years that involves adapting LLMs using "prompting."

In our current study, we investigate the performance of GPT LMs in this emerging paradigm and present preliminary results. Our research questions were: (1) How do LMs perform in detecting incoherent answers to typical math word problems among fourth-grade students? and (2) How does the performance of LMs compare to that of ML classifiers? We tested four different LLMs and found that they performed worse than the ML models, which had been trained on thousands of examples. In contrast, the LLMs were not specifically trained for this task and were only given a maximum of four examples. We also found that recursive questions posed the greatest difficulty for the GPT models. Another challenge was posed by the common misspellings among fourth-grade children.

We developed a specific prompt design for binary text classification using GPT models. Our prompt engineering approach involved manually creating intuitive templates through human inspection. Since the selection of prompts has a significant impact on results, we acknowledge that exploring alternative strategies can improve results. For example, we could use automated template learning, such as Prompt mining and Gradient-based search, proposed by Liu et al. (2023). Additionally, we could try methods that directly prompt the embedding space of the model, such as Prefix Tuning.

In our study, we map the output sentence of the LM to two labels, "coherent" and "incoherent," based on the presence of certain keywords in the output to create an effective predictive model. However, manual output mapping may result in suboptimal performance with LMs. Therefore, we are interested in exploring other strategies such as Prune-then-Search to find a better mapping, as well as continuous search for the answer space. Furthermore, we could explore in future research a new family of prompting methods based on multi-prompt learning, such as Prompt Ensembling (Liu et al., 2023).

To the best of our knowledge our results on the use of LLMs in detecting incoherent answers to open questions are new. However, we can compare our results with similar question-answer tasks in EDM that score student answers. One such task is ASAG, which has gained significant attention in education due to the growing number of both online and face-to-face students (Haller et al., 2022). According to Haller et al. (2022), recent advancements in NLP and ML have greatly influenced ASAG. For instance, Sung et al. (2019) demonstrated the efficacy of transformer-based pre-training on the SemEval-2013 benchmark dataset, resulting in an improvement of up to 10% in macro-average-F1 over state-of-the-art ML models.

Although large transformer-based LMs have shown promising results in solving ASAG tasks, they do not always outperform traditional models in specific domains. For instance, Gaddipati et al. (2020) evaluated the performance of four LLMs on the Mohler dataset and found poor results for GPT-2, GPT-1, and BERT. In fact, they reported that a traditional NLP approach using bag-of-words and term frequency-inverse document frequency outperformed GPT-2. While there is evidence of both positive and negative results with the new language architectures, including GPTs in EDM, there is currently no research on predicting incoherence in response to open-ended questions from fourth-grade students on an online mathematics platform. Our findings also suggest that LLMs have limitations in outperforming traditional ML models in predicting incoherence.

Our analysis on the performance of LLM models suggests that inferior performance may not be due to prompting, but rather an intrinsic limitation of the models. One such limitation is the ability to comprehend language, which is an important feature of LMs studied within NLP and natural language understanding (NLU). One of the canonical tasks in NLU is natural language inference (NLI) (Williams, A. et al., 2020). In NLI, a "premise" is given and the task involves figuring out if a "hypothesis" is true (entailment), false (contradiction), or uncertain (neutral). For example, given the premise "Juliet has 25 cookies and gives 5 to her friend", possible hypotheses are "Juliet has 20 cookies left" (entailment), "Juliet does not give cookies to her friend" (contradiction), and "Amanda has 5 cars" (neutral) (authors accessed on http://nlpprogress.com/english/natural_language_inference.html, March, 2023).

The generation of a diverse and competitive set to study NLI from pre-existing annotated datasets is a challenging task. To address this challenge, Demszky et al. (2018) proposed a new method to automatically derive NLI datasets from large-scale QA datasets. This approach suggests that any open-ended question-answer pair can be transformed into an example of NLI. Additionally, we propose that determining the incoherence in answers to open-ended questions is a sub-task of NLI. Verifying the incoherence of an answer (hypothesis) to an open-ended question (premise) is equivalent to verifying that the answer is coherent if it satisfies entailment or contradiction, and incoherent if it satisfies neutral.

In our experiments with LLM for incoherence detection, we observed that GPT models perform worse than ML models. Now this is not surprising, if we consider that incoherence detection is a sub-task of NLI. In fact, recent research by Brown et al. (2020) shows that GPT-3 does not perform significantly better than a model that randomly predicts on the ANLI dataset (Nie, Y. et al., 2020).

The limitations of GPT models may be similar to those observed in poor performance on NLI datasets. However, at the same time, our experiments indicate that GPT models face the greatest difficulty in questions that involve recursion. Interestingly, this feature is not exclusive to language but also observed in Computer Vision (CV).

Recursion is a significant challenge in various domains of AI, particularly in image classification. ML models often struggle with the classification of recursive images. Bongard problems (BPs) are a class of pattern recognition problems in the field of CV. They were initially introduced to evaluate human-level cognition for visual pure-intelligence pattern recognition (Bongard, M. M., 1970). Recursion is a key feature specific to BPs (e.g. BP No. 70), analogous to the feature present in Q3 type questions. The challenge of recursion is not limited to branches. Other shapes or patterns can exhibit recursion.

The complexity of recursion in visual patterns can be transferred to the shape of the object (e.g. BP No. 71) or a concept of numerosity (Depeweg et al., 2018; Youssef et al., 2022). However, addressing recursion in language remains unsolved. It is unclear how to modify the existing prompts to improve the ability of LLMs to predict incoherence in recursive questions. Nonetheless, knowledge transfer from visual reasoning could simplify the task of handling recursion in language and enhance the performance of LLMs. For instance, Huang,

S. et al. (2023) illustrate that multimodal-LLMs can benefit from knowledge transfer across language modalities and multimodal input. In conclusion, the challenge of recursion in AI remains a significant issue, and further research is needed to develop effective solutions.

On the other hand, the prevalence of spelling errors in the answers of fourth graders poses a challenge, as does the recursive nature of the questions. The occurrence of writing errors is a well-established phenomenon observed in both adults and K-12 students (Connors, R. J. et al., 1988). These errors manifest in various forms, ranging from simple spelling errors to incorrect capitalization. The frequency and types of spelling mistakes vary widely, with some being more common than others. Moreover, the prevalence of particular spelling errors may differ between native and non-native speakers of a language (Flor, M. et al., 2012). Generally, misspellings are the most frequently occurring type of error (Connors, R. J. et al., 1988), such as the misuse of diacritical marks or the omission of letters within words.

Spell-checking systems are available today to correct writing errors automatically, including misspellings. Despite their availability, these systems have limitations in accurately detecting and correcting errors (Ha, M. et al., 2016). Studies, such as Ha, M. et al. (2016), have evaluated how automated computerized scoring systems can be affected by spelling errors in open-ended assessments. The type of misspelling significantly impacts the performance of automatic systems in detecting errors. This issue is crucial because erroneous feedback to students may result from systems that make mistakes. Therefore, it is important to ensure the accuracy and effectiveness of automatic systems to provide appropriate feedback to students.

At present, few comprehensive studies have examined the impact of misspellings in fourth-grade students' open-ended responses on an online mathematics platform on automatic systems. Similarly, there is a lack of research on the impact of children's idiosyncratic spelling errors on automatic incoherence prediction. Our study suggests that misspellings by fourth-grade students have a negative and significant effect on automatic incoherence prediction when using LLMs such as GPT-3. We propose that this negative impact is primarily due to the fact that the training domain of LLMs captures text written by fourth graders to a lesser extent, or not at all, including the typical spelling mistakes made by students.

# Conclusion

Written argumentation is one of the basic skills in the mathematics curriculum. It helps students develop analytical and critical thinking skills, helps students learn to construct rigorous and logical arguments, and helps them develop communication skills. It is a basic tool for communicating their mathematical reasoning in an organized and effective manner. Moreover, its importance is increasing, as the automation of simpler processes will continue to advance in the coming years (Araya, 2021).

Written answers to open-ended questions could have a great effect on students' long-term learning. However, in order to ensure that we can harness the full power of open-ended questions, it is critical that the teacher immediately review the answers, and ask to redo those that are incoherent. Accessing real-time information of students' written explanations allows teachers to detect any negative attitudes and to give feedback to students who may be writing answers that are not only inaccurate but also incoherent. This is different from incorrect answers, as incoherent answers indicate students may not be engaged in the task and having negative attitudes.

However, real-time reviewing written responses to open-ended questions is very challenging. It can be very difficult for a teacher to provide real-time feedback to all their students in the classroom. It is a time-consuming and demanding task. Automating the detection of incoherent answers can help reduce the time it takes for the teacher to review. It is also important for teachers to review the peer reviews of their students' written responses, as this can provide valuable insight.

Moreover, an incoherence detector is a key component for the design of personalized intelligent tutors. They have to identify incoherent responses from students, and immediately ask to redo. It also has to inform the teacher of this early alert of negative attitudes. In our experience, 20% of responses from fourth-graders from vulnerable schools can be incoherent. This is different from incorrect answers. They could reveal a negative attitude from the student. Therefore, the strategy for handling incoherent answers is usually different from incorrect answers. With the correct detection of incoherent responses, intelligent tutors can provide a better learning experience for students.

Even though the performance in LLM is not very good in detecting incoherent responses, given its excellent discursive capacity, we propose to use it in a system that detects incoherence with ML but delivers the explanations through an LLM. They could constitute a powerful combination that supports the teacher explaining why the answer is incoherent. This combination could be very powerful to include as a conversational agent in an intelligent tutoring system. It could explain to the student why the answer is incoherent.

**Acknowledgments.** Support from ANID/PIA/Basal Funds for Centers of Excellence FB0003 is gratefully acknowledged.

# Declarations

## Funding

This work was supported by the Chilean National Agency for Research and Development (ANID), grant number ANID/PIA/Basal Funds for Centers of Excellence FB0003.

## Conflict of interest

The author declares no conflicts of interest. The funder had no role in the design of the study; in the collection, analyses, or interpretation of data; in the writing of the manuscript; or in the decision to publish the results.

## Ethical approval

Ethical review and approval were waived for this study, due to it being a class session during school time. The activity was revised and authorized by the respective teachers.

## Consent to participate

Student consent was waived due to authorization from teachers. Given that there are no patients but only students in a normal session in their schools, within school hours, and using a platform that records their responses anonymously, the teachers authorized the use of anonymized information.

## Availability of data and materials

Not available.

## Code availability

Code available in public GitHub repository: https://github.com/furrutiav/coherence-prediction.

## Authors' contributions

Conceptualization, R.A. and F.U.; methodology, R.A. and F.U.; software, F.U.; validation, R.A. and F.U.; formal analysis, R.A. and F.U.; investigation, R.A. and F.U.; resources, R.A.; data curation, F.U.; writing—original draft preparation, F.U.; writing—review and editing, R.A. and F.U.; visualization, F.U.; supervision, R.A.; project administration, R.A.; funding acquisition, R.A. All authors have read and agreed to the published version of the manuscript.

# Appendix

## Generative Pre-trained Transformers

In recent years, LLMs have been developed, such as GPT-3, BLOOM, YouChat, and ChatGPT, which are based on the state-of-the-art GPT architecture and are capable of generating coherent text on a wide range of topics and styles (Brown et al., 2020; Muennighoff et al., 2022; Ouyang et al., 2022; Stiennon et al., 2020). While GPT-3 and BLOOM are general-purpose models, YouChat and ChatGPT are specifically designed for online conversations, and they employ natural language technologies trained using self-supervised learning on extensive corpora to provide real-time responses to user queries. YouChat utilizes search engines to generate responses, while ChatGPT focuses on producing personalized responses by leveraging human-feedback training (Stiennon et al., 2020).

These models are increasingly being adopted in various applications, including general text generation and response automation in online conversations (Kasneci et al., 2023). GPT-3 represents a highly advanced NLP system developed by OpenAI, which uses a self-attention transformer neural network to generate text from large amounts of pre-training on textual data (Brown et al., 2020). With greater processing power and a greater number of parameters, GPT-3 exhibits the ability to generate text that is nearly indistinguishable from human-written text, as well as to complete sentences effectively. However, as noted by Dale (2021), language models such as GPT-3 may produce nonsensical outputs as they become longer, and they may also replicate any biases contained in the underlying training data, leading to potentially false statements.

BLOOM is a scalable natural language model developed by BigScience, a research group focused on developing advanced natural language models and promoting open and collaborative research in this area (Scao et al., 2022). BLOOM-large is a larger version of the model consisting of approximately 176B parameters and has been trained on a large quantity of cross-lingual data. Additionally, BLOOMz is an enhanced version of BLOOM that uses Multitasking Prompted Fine-tuning (MTF) to enable the model to undertake new tasks in a multilingual setting without prior training (Muennighoff et al., 2022).

ChatGPT is a natural language model tailored to generate responses in chat conversations, utilizing a self-attention architecture and trained on large-scale textual datasets (Ouyang et al., 2022). Distinguished from other models, ChatGPT is designed explicitly for deployment in interactive response generation for chat conversations and virtual assistants. YouChat is an automated chatbot that employs natural language technologies, including the GPT-3.5 model developed by OpenAI, to generate instantaneous responses to user inquiries. Alongside ChatGPT, YouChat utilizes Google search for general outcomes and Microsoft Bing for specific searches, such as code snippets (authors accessed on [https://you.com/](https://you.com/), February 2023).

## Prompt-based Training Strategies

Prompting is a learning technique that presents a partial sentence to a LLMs model to improve its performance (Liu, P. et al, 2023). Prompt-based training can improve the quality of model outputs by providing the model with an idea of what to generate, and it is used in a wide range of NLP tasks, such as Question Answering (QA) and Summarization (Radford, A. et al., 2019). Prompt-based training strategies, including zero-, one-, and few-shot (Figure 2).

As an illustration, one can utilize prompting to train a LM to recognize incoherence in answers to open-ended questions. Specifically, the model can be given a sentence fragment beginning with the phrase "If an answer to an open-ended question can be either coherent or incoherent, then the answer R to the question P is", where R represents an answer, and P represents a given question. By completing the sentence fragment, the model is expected to provide a classification of whether the answer is coherent or incoherent to the open-ended question. This prompting strategy is known as zero-shot learning.

Zero-shot refers to the model's ability to generate text for tasks that were not included in its original training set by using prompts that provide information about the task and its context (Brown, T. et al., 2020). One-shot involves training the model with a single example for a specific task, while few-shot involves training the model with a small set of examples. These strategies are particularly useful when little training data is available or when the model's ability to generate more accurate responses needs to be improved (Liu, P. et al, 2023).

*Figure 2. (**Left**) Structure of the Prompt-based Training Strategies. (**Right**) Input to the GPT and its two outputs. (Template originally in Spanish).*

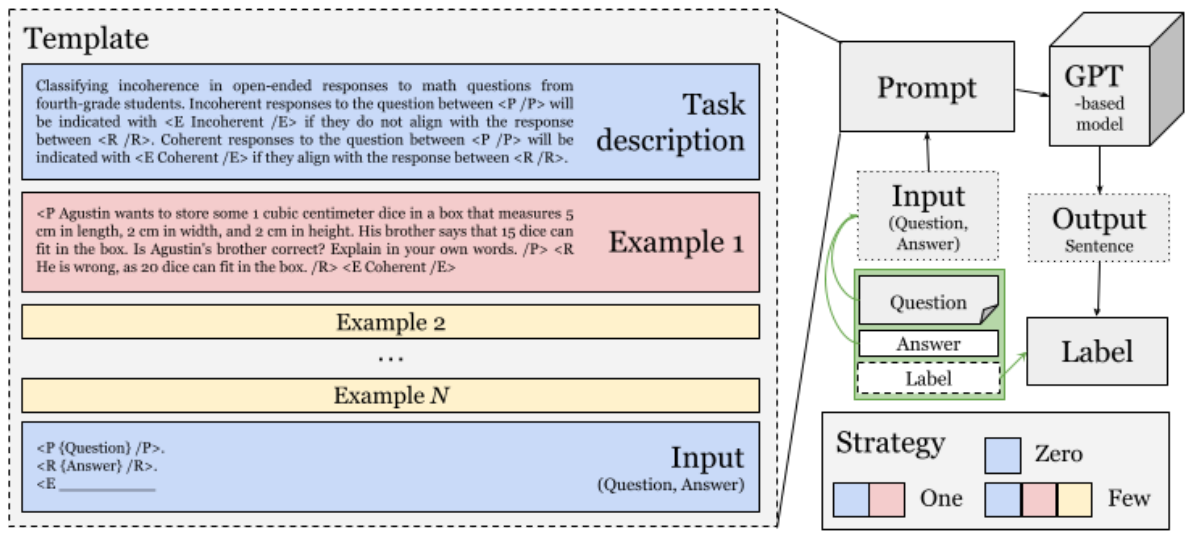

# Prompts

*Prompt YouChat. Zero-shot.(Prompts originally in Spanish)*

| Input | Hello, are you ready? |
|---|---|
| YouChat | Hello! Yes, I'm ready to help you with any questions you have! How can I assist you? |
| Input | You are an incoherence classifier for open-ended questions from fourth-grade math students. I will give you a question and different answers. Then you have to say whether the answer is coherent or incoherent with the question, along with an explanation. Coherent answers may be incorrect. Are you ready? |
| YouChat | Yes, I'm ready! To classify the coherence of responses to open-ended questions from fourth-grade math students, I will read the question and evaluate each response individually to determine whether it is coherent or not. If an answer is coherent but incorrect, then it will also be considered incoherent. To explain my evaluation, I will provide a detailed explanation of why I consider the answer to be coherent or incoherent with the question. Do you have any particular questions that you would like me to classify for coherence? |
| Input (data to be labeled) | [Coherent or incoherent?] <Question> {q}}; <Answer> {{a}}. |
| Input (example) | [Coherent or incoherent?] <Question> What is the difference between 345 and 445?; <Answer> The difference is 100. |
| YouChat (example) | This answer is coherent with the question. The question was about the difference between 345 and 445, and the answer is correct in stating that the difference is 100. |

*Prompts GPT-3. (Prompts originally in Spanish)*

| Input (zero-shot) | Hello, are you ready? You are an incoherence classifier for open-ended questions from fourth-grade math students. I will give you a question and different answers. Then, you have to determine if the answer is coherent or incoherent to the question. We will mark the questions between <P {question} /P> and the answers with <R {answer} /R>. You will provide the label incoherent or coherent between <E {label} /E>. Are you ready? <P>{{q}}/P>. <R>{{a}}/R>. <E |
|---|---|
| (zero-shot+ | Hello, are you ready? You are an incoherence classifier for open-ended math questions from fourth-grade students. I will |
| adv+i) | give you a question and different answers, and you have to determine whether the answer is coherent or incoherent to the question. We will mark questions with <P {{question}} /P> and answers with <R {{answer}} /R>. You will provide the label "incoherent" or "coherent" between <E {{label}} /E>. Please take extra care in the following cases: Recursivity: If we ask for P, but P has multiple questions P(Q1, Q2, etc.), then you must distinguish between P and the Qs. Correctness: Incorrect answers can still be coherent. To better detect coherence, consider the following: (1) For questions that require computation, a coherent answer is one that includes at least one appropriate number. (2) For questions that require explanation, justification, or argumentation, a coherent answer is one that includes an explanation, justification, or argument. (3) For questions that require determining who is correct or if a statement is true, a coherent answer is one that responds affirmatively and/or includes the name of a character from the question. Are you ready? <P {q} /P>. <R {a} /R>. <E |
| (zero-shot+ adv) | Hello, are you ready? You are an incoherence classifier for open-ended questions from fourth-grade math students. I will give you a question and different answers. Then, you have to determine if the answer is coherent or incoherent with the question. We will mark the questions between <P {question} /P> and the answers with <R {answer} /R>. You will deliver the label "incoherent" or "coherent" between <E {label} /E>. Take it slower in the following cases: Recursivity: If we ask about P, but P has several questions P(Q1, Q2, etc), then you must distinguish between P and the Qs. Correctness: Incorrect answers can still be coherent. Are you ready? <P {{q}} /P>. <R {{a}} /R>. <E |
| (one-shot+a dv) | Hello, are you ready? You are an incoherence classifier for open-ended questions from fourth-grade math students. I will give you a question and different answers, and you have to determine if the answer is coherent or incoherent to the question. We will mark the questions with <P {{question}} /P> and the answers with <R {{answer}} /R>. You will provide the label "coherent" or "incoherent" between <E {{label}} /E>. Here is an example: <P Agustin wants to store some 1-cubic-centimeter dice in a box that measures 5 cm long, 2 cm wide, and 2 cm high. His brother says that 15 dice can fit in the box. Is Agustin's brother correct? Explain in your own words. /P> <R It's wrong, 20 dice can fit in the box. /R> <E Coherent /E> Please slow down in the following cases: Recursion: If we ask about P, but P has multiple questions P(Q1, Q2, etc.), then you need to distinguish between P and Q. Correctness: Incorrect answers can still be coherent. Are you ready? <P {q} /P>. <R {a} /R>. <E |
| (two-shot+a dv) | Hello, are you ready? You are an incoherence classifier for open-ended questions from fourth-grade math students. I will give you a question and different answers. Then, you have to determine whether the answer is coherent or incoherent to the question. We will mark the questions with <P {{question}} /P> and the answers with <R {{answer}} /R>. You will deliver the label "incoherent" or "coherent" between <E {{label}} /E>. I will give you two examples: <P Agustin wants to store some 1 cm3 dice in a box that measures 5 cm in length, 2 cm in width, and 2 cm in height. His brother says that 15 dice can fit in the box. Is what Agustin's brother saying correct? Explain in your own words. /P> <R It's wrong, 20 dice can fit. /R> <E Coherent /E> <P Use the digits 1420 and 1356 to create a subtraction problem. /P> <R My dad gave me 1420 pesos to buy chocolate, and it costs 1356 pesos. How much money do I have left? /R> <E Coherent /E> Please take it slower in the following cases: Recursivity: If we ask about P, but P has several questions P(Q1, Q2, etc.), then you must distinguish between P and the Qs. Correctness: Incorrect answers can still be coherent. Are you ready? <P {q} /P>. <R {a} /R>. <E |
| (three-shot+ adv) | Hello, are you ready? You are an incoherence classifier for open-ended questions from fourth-grade math students. I will give you a question and different answers. Then you have to determine whether the answer is coherent or incoherent with the question. We will mark the questions with <P {question} /P> and the answers with <R {answer} /R>. You will provide the coherent or incoherent tag within <E {tag} /E>. I will give you two examples: <P Agustín wants to store dice that measure 1 cubic cm in a box that is 5 cm long, 2 cm wide, and 2 cm high. His brother says that 15 dice can fit in the box. Is what Agustin's brother saying correct? Explain in your own words. /P> <R It's wrong, 20 dice fit in the box. /R> <E Coherent /E> <P Use the digits 1420 and 1356 to create a subtraction problem. /P> <R My dad gave me 1420 pesos to buy chocolate, and it costs 1356 pesos. How much money do I have left? /R> <E Coherent /E> Please take it slower in the following cases: Recursion: If we ask about P, but P has multiple questions P(Q1, Q2, etc), then you must distinguish between P and the Q's. Correctness: Incorrect answers can still be coherent. Are you ready? <P {q} /P>. <R {a} /R>. <E |
| (four-shot+a dv) | Hello, are you ready? You are an incoherence classifier in responses to open questions from fourth-grade math students. I will give you a question and different answers, and then you have to say whether the answer is coherent or incoherent to the question. We will mark the questions between <P {{question}} /P> and the answers with <R {{answer}} /R>. You will deliver the label incoherent or coherent between <E {{label}} /E>. I will give you two examples: <P Agustin wants to store 1 cubic cm dice in a box that measures 5 cm long, 2 cm wide, and 2 cm high. His brother says that 15 dice can fit in the box. Is what Agustin's brother said correct? Explain in your own words. /P> <R It's wrong, it can fit 20 dice /R> <E Coherent /E> <P Create a subtraction problem with the digits 1420 and 1356. /P> <R My dad gave me 1420 pesos to buy a chocolate, and it costs 1356 pesos. How much money do I have left? /R> <E Coherent /E> <P Amalia bought 1.75 cm of elastic, and Antonieta bought 1.57 cm. A friend of theirs told them that they bought a total of 3 meters of elastic. Is what their friend saying correct? Explain in your own words. /P> <R I think it's wrong because it's 3.32 /R> <E Incoherent /E> <P What did you learn in Friday's activity? /P> <R I learned that I can add with animals, and we can learn more. /R> <E Coherent /E> Please go slower in the following cases: Recursion: If we ask for P, but P has several questions P(Q1, Q2, etc.), then you must distinguish between P and the Qs. Correctness: Incorrect answers can be coherent. Are you ready? <P {q} /P>. <R {a} /R>. <E. |

*Prompts BLOOMz. (Prompts originally in Spanish)*

| | |
|---|---|
| Input (one-shot) | Classifying incoherence in open-ended responses to fourth-grade math students' questions. It will be indicated with <E Incoherent /E> if the response between <R /R> is incoherent with the question between <P /P>. It will be indicated with <E Coherent /E> if the response between <R /R> is coherent with the question between <P /P>. <P Agustin wants to store some 1 cubic centimeter dice in a box that measures 5 cm in length, 2 cm in width, and 2 cm in height. His brother says that |
| | 15 dice can fit in the box. Is what Agustin's brother says correct? Explain in your own words. /P> <R It's wrong because 20 dice can fit. /R> <E Coherent /E> <P {q} /P>. <R {a} /R>. <E Incoherent /E> |
| two-shot | Classify incoherence in responses to open questions from fourth-grade math students. It will be indicated with <E Incoherent /E> if the response between <R /R> is incoherent with the question between <P /P>. It will be indicated with <E Coherent /E> if the response between <R /R> is coherent with the question between <P /P>. <P Agustin wants to store some 1 cubic centimeter dice in a box that measures 5 cm in length, 2 cm in width, and 2 cm in height. His brother says that 15 dice can fit in the box. Is what Agustin's brother says correct? Explain in your own words. /P> <R It's wrong, it can fit 20 dice /R> <E Coherent /E> <P Maria and her husband cooked a tortilla yesterday, divided it into 6 equal parts. Maria ate 2/6 and her husband ate 3/6. What fraction of the tortilla is left? /P> <R There was a piece of tortilla left and the last one was "a" and the tortilla /R> <E Coherent /E> <P {q} /P>. <R {a} /R>. <E Incoherent /E> |
| three-shot | Classifying incoherence in open-ended math questions from fourth grade students. It will be indicated with <E Incoherent /E> if the response between <R /R> is incoherent to the question between <P /P>. It will be indicated with <E Coherent /E> if the response between <R /R> is coherent to the question between <P /P>. <P Agustin wants to store some 1 cubic centimeter dice in a box that measures 5 cm long, 2 cm wide, and 2 cm high. His brother says that 15 dice can fit in the box. Is what Agustin's brother is saying correct? Explain in your own words. /P> <R It is wrong because 20 dice can fit. /R> <E Coherent /E> <P Maria and her husband cooked a tortilla yesterday, they divided it into 6 equal parts. Maria ate 2/6 and her husband ate 3/6. What fraction of the tortilla is left? /P> <R There was a piece of tortilla left and the last one was "a" and the tortilla. /R> <E Coherent /E> <P Claudio has 12 pencils to distribute equally in 3 pencil cases. To do this, his friend Paula tells him to put 6 pencils in each case, and then he won't have any left over. Is Claudio's friend correct? (Explain the answer in your own words) /P> <R Paula is not correct because she is asking for more pencil cases. /R> <E Coherent /E> <P {q} /P>. <R {a} /R>. <E |
| four-shot | Classifying incoherence in responses to open-ended math questions from fourth-grade students. Responses that are incoherent with the question will be indicated by <E Incoherent /E>, while responses that are coherent with the question will be indicated by <E Coherent /E>. <P Agustin wants to store some 1-cubic-centimeter dice in a box that is 5 cm long, 2 cm wide, and 2 cm high. His brother says that 15 dice can fit in the box. Is Agustin's brother correct? Explain in your own words. /P> <R It's wrong because 20 dice can fit. /R> <E Coherent /E> <P Maria and her husband cooked a tortilla yesterday, dividing it into 6 equal parts. Maria ate 2/6 and her husband ate 3/6. What fraction of the tortilla remains? /P> <R There was a piece of tortilla left, and the last one was 'a' and the tortilla. /R> <E Coherent /E> <P Claudio has 12 pencils to distribute equally among 3 pencil cases. To do so, his friend Paula tells him to put 6 pencils in each case so that he won't have any left over. Is Claudio's friend correct? (Explain your answer in your own words) /P> <R Paula is not correct because she's asking for more cases. /R> <E Coherent /E> <P Francisco and Camila have a collection of action figures from their favorite series. Francisco has 4/8 of the collection and Camila has 2/8. If they lose 5/8 of the total number of figures they have, how many figures are left? (Write your answer appropriately) /P> <R IT'S GOOD /R> <E Incoherent /E> <P {q} /P>. <R {a} /R>. <E |

*Prompts BLOOM-large. (Prompts originally in Spanish)*

| | |
|---|---|
| Input (zero-shot) | This is an incoherence classifier in responses to open-ended questions from fourth-grade math students. We will provide questions and answers, and then determine if each response is coherent or incoherent to its corresponding question. We will mark the questions between <P and /P>, and the answers between <R {{response}} /R>. Incoherent labels will be given by <E Incoherent /E>, while coherent ones will be marked as <E Coherent /E>, as appropriate. We will always provide a label and end the page with /E>. <P {q} /P> <R {a} /R> <E |
| zero-shot+adv | This is an incoherence classifier for open-ended questions from fourth-grade math students. We will provide questions and answers, and then determine whether each answer is coherent or incoherent to its question. We will mark the questions between <P and /P>, and the answers between <R {{answer}} /R>. Incoherent labels will be given by <E Incoherent /E>, and coherent labels by <E Coherent /E>, as appropriate. We will always provide a label and end the page with /E>. We will go slower in the following cases: Recursion: If we ask about P, but P has multiple questions P(Q1, Q2, etc.), then you must distinguish between P and the Qs. Correctness: Incorrect answers may be coherent. <P {q} /P>. <R {a} /R>. <E |

*Prompt GPT-3 Q3. (Prompts originally in Spanish)*

| | |
|---|---|
| Input (zero-shot) | Question: {q}. Felipe's answer: {a}. Answer with yes or no to 3 different queries: Is the answer correct within the question? Is Felipe's answer correct? Is Felipe's answer coherent? |

# Answers ChatGPT

*Example 1. "D is a ten". (Example original in Spanish)*

| | |
|---|---|
| Input (id: 506) | Question: D is a ten, why is 2D apples the same as 20 apples? Explain. Answer: Because the desenavale is 10. Is the answer to the question coherent? |

| ChatGPT | The answer is incoherent and does not make sense. Stating that "desenavale" is 10 is not a logical explanation for the equivalence of 2D apples with 20 apples. The correct answer is that "D" represents the quantity of tens in a number. Therefore, 2D apples means 2 times the quantity of tens, which translates to 20 apples. |
|---|---|

*Example 3. "Farmer Braulio". (Example original in Spanish)*

| Input (id: 201) | Question: Farmer Braulio distributes his harvest of 35 corn cobs in 7 trucks. A worker said that they will put 4 corn cobs in each truck. Is the worker correct? Explain in your own words. Answer: It is incorrect because 7 trucks × 5 is 35 corn cobs. Is the answer coherent with the question? |
|---|---|
| ChatGPT | The answer is not coherent with the question as it has provided an opposite answer to the original question. Instead of explaining whether the worker is correct in saying that each truck will have 4 corn cobs, an answer has been provided that indicates that 7 trucks multiplied by 5 are 35 corn cobs. Therefore, an answer that directly addresses the original question is needed to determine whether the worker is correct or not. |

*Example 4. "Amalia's friend and Antonieta".*

| Input (id: 84233 /4496) | Question: Amalia bought 1.75 cm of elastic and Antonieta 1.57 cm. A friend of theirs told them that they bought a total of 3 meters of elastic. Is what their friend telling them correct? Explain in your own words. Answer: I think it's wrong because it's 03.32 cm. Is the answer coherent with the question? |
|---|---|
| ChatGPT | The answer is not coherent with the question because it indicates a quantity in centimeters instead of meters, and it does not explain whether 3 meters is the total amount purchased by both people or if it is the amount purchased by each of them. In order to determine whether the answer is correct or not, this information would need to be clarified and the corresponding unit conversion performed. |

# Who's the Best Detective? LLMs vs. MLs in Detecting Incoherent Fourth Grade Math Answers

Felipe Urrutia*† and Roberto Araya† *Corresponding author(s). E-mail(s): furrutia@dim.uchile.cl; Contributing authors: roberto.araya.schulz@gmail.com; †These authors contributed equally to this work.

Felipe Urrutia*† and Roberto Araya† *Corresponding author(s). E-mail(s): furrutia@dim.uchile.cl; Contributing authors: roberto.araya.schulz@gmail.com; †These authors contributed equally to this work.

**Acknowledgments** Support from ANID/PIA/Basal Funds for Centers of Excellence FB0003 is gratefully acknowledged. **Funding** This work was supported by the Chilean National Agency for Research and Development (ANID), grant number ANID/PIA/Basal Funds for Centers of Excellence FB0003. **Conflict of interest** The author declares no conflicts of interest. The funder had no role in the design of the study; in the collection, analyses, or interpretation of data; in the writing of the manuscript; or in the decision to publish the results. **Ethical approval** Ethical review and approval were waived for this study, due to it being a class session during school time. The activity was revised and authorized by the respective teachers. **Consent to participate** Student consent was waived due to authorization from teachers. Given that there are no patients but only students in a normal session in their schools, within school hours, and using a platform that records their responses anonymously, the teachers authorized the use of anonymized information.

**Availability of data and materials** Not available. **Code availability** Code available in public GitHub repository: https://github.com/furrutiav/coherence-prediction. **Authors' contributions** Conceptualization, R.A. and F.U.; methodology, R.A. and F.U.; software, F.U.; validation, R.A. and F.U.; formal analysis, R.A. and F.U.; investigation, R.A. and F.U.; resources, R.A.; data curation, F.U.; writing—original draft preparation, F.U.; writing—review and editing, R.A. and F.U.; visualization, F.U.; supervision, R.A.; project administration, R.A.; funding acquisition, R.A. All authors have read and agreed to the published version of the manuscript.

**Roberto Araya** (roberto.araya.schulz@gmail.com; ORCID 0000-0003-2598-8994) is a Full Professor at the Institute of Education, University of Chile. He is a Ph.D. in Electrical Engineering from the University of California at Los Angeles. He obtained the Best Paper Award of the 10th International Conference in Methodologies and Intelligent Systems for Technology Enhanced Learning MIS4TEL 2020, Italy; Best Paper Award of 10th International Conference on Web-based Learning - ICWL 2011, Hong-Kong; Best Poster Award of 12th European Conference on Technology Enhanced Learning - ECTEL 2017, Estonia. He was also nominee for the Best Paper Award of the Artificial Intelligence in Education Conference AIED 2022, Durham, UK. He has leaded projects funded by the International Development Research Centre (IDRC) Canada, the Finnish Academy of Sciences together with ANID Chile, the Inter-American Development Bank, and the Education Endowment Foundation.

**Felipe Urrutia** (furrutia@dim.uchile.cl; ORCID 0000-0003-0809-5334) is a candidate for a Master's degree in Computer Science at the University of Chile. He has a degree in Engineering Sciences, mention in Mathematics, from the Faculty of Physical and Mathematical Sciences of the University of Chile. He is an assistant researcher at Center for Advanced Research in Education, University of Chile, Chile, and Center for Research for Learning and Teaching, University of Jyväskylä, Finland. He has published at least one article in a peer-reviewed journal in Education indexed in WoS.